\documentclass[10pt,twocolumn,letterpaper]{article}

\usepackage{cvpr}              % To produce the CAMERA-READY version
\definecolor{cvprblue}{rgb}{0.21,0.49,0.74}
\usepackage[pagebackref,breaklinks,colorlinks,allcolors=cvprblue]{hyperref}
\usepackage{tabularx}

\def\paperID{} % *** Enter the Paper ID here
\def\confName{3D-LLM/VLA}
\def\confYear{2026}

\title{Boosting MLLM Spatial Reasoning with Geometrically Referenced 3D Scene Representations}

\author{Jiangye Yuan \quad Gowri Kumar \quad Baoyuan Wang \\
Zillow Group\\
}

\begin{document}
\maketitle
\begin{abstract}
While Multimodal Large Language Models (MLLMs) have achieved remarkable success in 2D visual understanding, their ability to reason about 3D space remains limited. To address this gap, we introduce geometrically referenced 3D scene representations (GR3D). Given a set of input images, GR3D annotates objects in the images with unique IDs and encodes their 3D geometric attributes as textual references indexed by these IDs. This representation enables MLLMs to interpret 3D cues using their advanced language-based skills in mathematical reasoning, while concurrently analyzing 2D visual features in a tightly coupled way. We present a simple yet effective approach based on GR3D, which requires no additional training and is readily applicable to different MLLMs. Implemented in a zero-shot setting, our approach yields substantial improvements on challenging spatial reasoning benchmarks, boosting \mbox{GPT-5} performance by 9\% on VSI-Bench and 12\% on MindCube. Qualitative studies further demonstrate that GR3D empowers MLLMs to perform complex spatial reasoning with highly sparse input views.

\end{abstract}    
\section{Introduction}
\label{sec:intro}

Understanding 3D environments is a fundamental aspect of human cognition. Developing models capable of interpreting such environments and interacting through natural language has become a central objective in the 3D scene understanding community~\cite{huang2024embodied, ma2024llms}. Multimodal Large Language Models (MLLMs), trained with massive amounts of vision and language data, have achieved remarkable success in 2D visual understanding~\cite{team2023gemini, hurst2024gpt, qwen2vl_dynamic_resolution, chen2024internvl}. Motivated by the success, they have been applied to 3D tasks in the hope that their 2D knowledge can be generalized to 3D cases. However, benchmarking studies reveal that MLLMs perform far below human level and struggle with tasks humans accomplish with ease~\cite{yang2025thinking}.  

Since MLLMs do not natively process 3D data, a complementary line of work extends vision-language models with explicit 3D inputs, such as point clouds, as an additional modality. Despite steady progress, these fine-tuned models do not show a clear advantage over leading MLLMs on spatial reasoning benchmarks~\cite{jia2025omnispatial, zhang2025pointvisiontext}. A key bottleneck lies in data availability. Since 3D scene-language datasets are orders of magnitude smaller than internet-scale vision-language corpora, fine-tuning is typically restricted to relatively small models with limited reasoning and language interaction capabilities. Furthermore, it remains a fundamental challenge to design algorithms that enable models to effectively utilize and understand 3D data. As shown in recent studies~\cite{zhang2025pointvisiontext}, models fine-tuned with 3D data exhibit shallow spatial understanding, frequently failing on simple reversed reasoning tasks (e.g., if A is above B, determining what is below A).   

\begin{figure*}[t]
  \centering
  \includegraphics[width=0.95\textwidth]{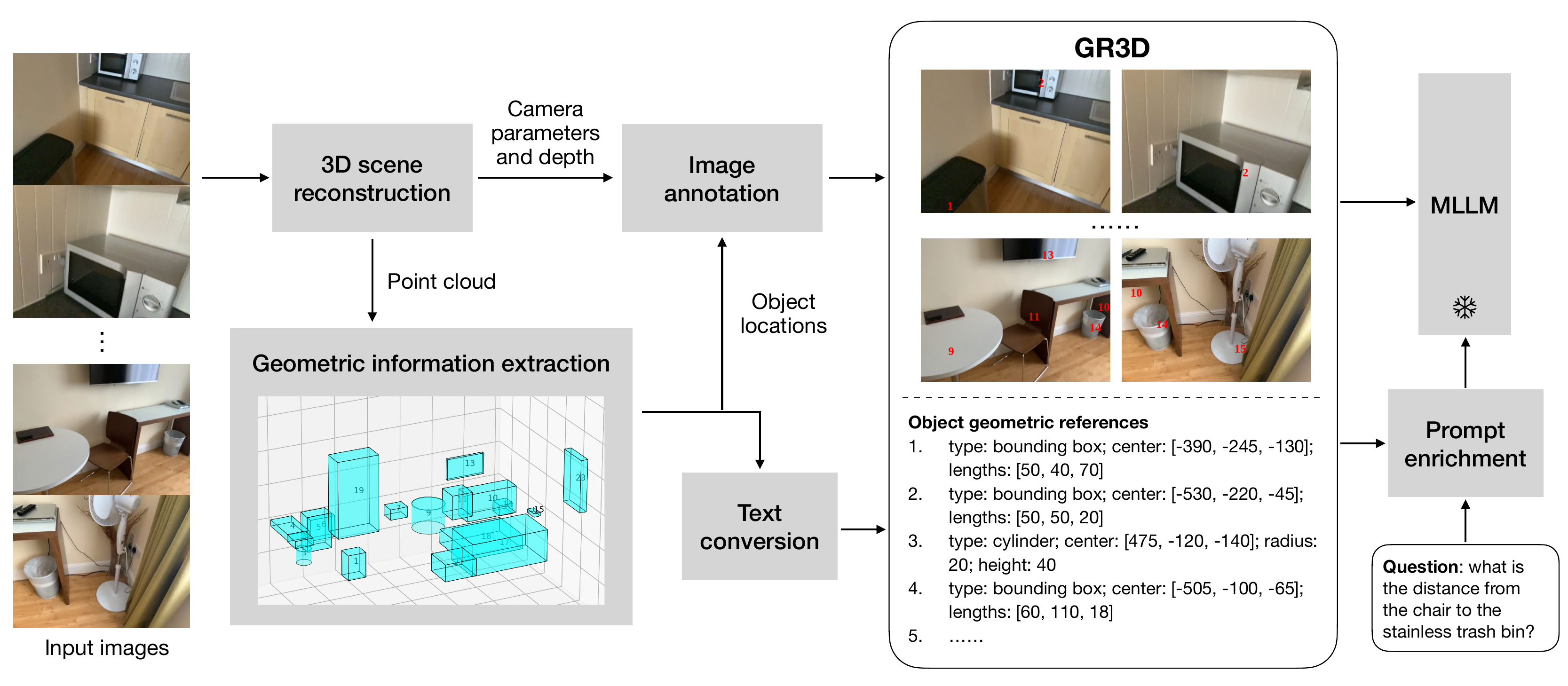}
  \caption{An overview of GR3D framework. Given a collection of images, our method reconstructs 3D scenes, extracts object-level geometric attributes, and transforms them into a GR3D representation, which consists of annotated images and textual references of geometric attributes. Such paired text and images are provided to a MLLM to perform spatial reasoning tasks.}
  \label{fig:3drep}
 \end{figure*}

Recent work has analyzed MLLMs' performance on spatial understanding tasks and examined their internal representations of space~\cite{yang2025thinking}. The findings indicate that MLLMs have difficulty reasoning about spatial relationship, such as distance, direction, and object size, and lack a holistic, global scene understanding. To address these limitations, we propose geometrically referenced 3D scene representations (GR3D). In this representation, objects are annotated in images with their unique IDs, and a list of textual references encode object geometries in a global 3D coordinate space. An example is shown in~\cref{fig:3drep}. It can be seen that text and images are explicitly cross-referenced via object IDs. Integrated with this representation, MLLMs can exploit their strong language-based abilities in mathematical and geometric reasoning~\cite{frieder2023mathematical, pan2025geologic}, while also flexibly accessing 2D semantic cues. This leads to precise, grounded, and interpretable spatial reasoning.  

An overall framework of leveraging GR3D for spatial reasoning is illustrated in~\cref{fig:3drep}. GR3D generation builds on recent advances in neural 3D reconstruction. Methods such as VGGT~\cite{wang2025vggt} and $\pi^3$~\cite{wang2025pi} can reconstruct 3D scenes from uncalibrated, unposed images with high accuracy and consistency. From reconstructed scenes, we derive object geometric attributes (\eg., bounding boxes, primitive types, and shape parameters) and convert them into structured textual references indexed by object IDs. Using estimated camera parameters, we project object centers into images and annotate projected locations with corresponding object IDs. The annotated images replace the original images as visual input for MLLMs, while textual references are combined with task-specific instructions to form text prompts. Our method does not require training and can work along with various MLLMs. 

We evaluate our approach on two challenging spatial reasoning benchmarks, the Visual-Spatial Intelligence Benchmark (VSI-Bench)~\cite{yang2025thinking} and MindCube~\cite{yin2025spatial}. We implement our method in a zero-shot manner. We neither fine-tune on any data that overlap with this benchmark dataset nor construct similar question-answer pairs for training. Nevertheless, our method achieves state-of-the-art benchmarking performance, demonstrating strong generalization and effectiveness. Ablation studies further confirm the critical role of GR3D in achieving this performance.   

Our contributions are summarized as follows:
\begin{itemize}
    \item We introduce GR3D, a new 3D scene representation that allows MLLMs to access both 2D visual cues and 3D geometric information in a tightly linked fashion. 
    \item We present an effective approach that leverages GR3D to enhance the spatial reasoning capabilities of MLLMs. 
    \item Experimental results show that our approach significantly outperforms previous state-of-the-art models on spatial reasoning tasks in a zero-shot setting. We also showcase that GR3D enables effective spatial reasoning from very sparse input views, where even leading MLLMs such as GPT-5 tend to fail. 
\end{itemize}

\section{Related Work}

Multimodal models seek to integrate vision, language, and other modalities within a unified framework. Early work, such as CLIP~\cite{radford2021learning} and ALIGN~\cite{jia2021scaling}, exploits contrastive learning to align image and text representations, laying the foundation for cross-modal reasoning. Subsequent models, like Flamingo ~\cite{alayrac2022flamingo} and BLIP-2 ~\cite{li2023blip}, are able to take visual and linguistic input and perform various text generation tasks with great zero-shot generalization. More recently, proprietary models including Gemini~\cite{team2024gemini} and GPT-5~\cite{hurst2024gpt}, as well as open-source alternatives (e.g., Qwen-VL~\cite{bai2023qwen}, InternVL~\cite{chen2024internvl}, and LLaVA~\cite{liu2023llava}), have demonstrated even greater capabilities in sophisticated reasoning and interactive human-AI dialogues. 

To enhance the 3D understanding capabilities of MLLMs, prior work has explored various 3D scene representations, including point cloud features ~\cite{Chen2024LL3DA, Xu2024PointLLM}, multi-view image embeddings lifted into 3D space ~\cite{SceneLLM2024, 3DLLM2023, LLaVA3D2024, Zheng2025Video3DLLM}, object-centric formulations~\cite{wang2023chat, huang2024chat, huang2024embodied}. However, most of these approaches require fine-tuning for models to understand the representations, and thus suffer from the aforementioned issues of data scarcity and learning effectiveness. SpatialPIN~\cite{Ma2024SpatialPIN} estimates depth from an input image and extracts text-based 3D priors, which are used to interact with MLLMs. Although this approach operates without training, it is difficult for MLLMs to associate objects 3D priors with visual features, especially when dealing with complex scenes captured by multiple images. GPT4Scene~\cite{Qi2025GPT4Scene} proposes complementing input images with a Bird's Eye View (BEV) synthesized via 3D reconstruction. While BEV encodes global information, it is non-trivial to select a viewpoint that ensures visibility of all objects, and MLLMs still require fine-tuning to better understand such BEV images. 

In contrast, our GR3D approach provides a 3D scene representation that can be natively understood by MLLMs without any fine-tuning. The representation tightly couples visual features with 3D geometric information, which enables MLLMs to reason in a more precise and coherent way. Our approach significantly enhances MLLM performance on spatial reasoning tasks in complex scenes. 
\section{Method}

In this section, we present the approach to leveraging GR3D for boosting MLLM spatial reasoning. In~\cref{subsec:recnstr}, we describe the method to extract 3D geometric information from input images. Then, we discuss the use of textual references to represent geometric information in~\cref{subsec:lan}. Finally, we elaborate on the method to associate images with textual references in~\cref{subsec:im}. 

\subsection{3D Scene Analysis}
\label{subsec:recnstr}

Given a set of multi-view images, we employ neural 3D reconstruction models to recover 3D scenes and camera poses. Unlike traditional Structure-from-Motion (SfM) pipelines that rely on handcrafted keypoint matching, such models use dense feature correlation and neural optimization, yielding reconstructions that are highly robust under textureless regions, occlusions, and wide baseline conditions. The output consists of a dense depth map $D$ for each input image along with the corresponding Camera intrinsics $K$ and extrinsics $(R, t)$. A 3D point cloud in a global coordinate system can be derived by unprojecting pixels with their depth values.

To extract object-level representations, we segment the point cloud into clusters corresponding to candidate objects. Two strategies are commonly used, depending on the point cloud quality. One strategy is to run 2D semantic segmentation, back-project labels onto 3D points, and fuse labels across multiple views. This approach benefits from mature image segmentation models and remains effective when the 3D point cloud is noisy. The other strategy is to apply neural networks trained directly on point clouds~\cite{qi2017pointnet++, schult2023mask3d}. These methods produce more accurate results when dense, high-quality point clouds are available, though performance may vary across datasets.  

For each object cluster, we estimate geometric attributes that range from coarse extents to detailed structures. Bounding boxes provide a compact representation of object extents defined by their centers, orientations, and side lengths. In many applications, such representations provide an adequate approximation for geometric inference and reasoning. When finer structural details are required, we fit analytic primitives such as cuboids, cylinders, spheres, and prisms. The parameters of these primitives are estimated by minimizing the residual distance between cluster points and the corresponding primitive surface. To handle noise and outliers, RANSAC-based fitting~\cite{schnabel2007efficient} and learning-based approaches~\cite{zou20173d, li2019supervised} can be applied for further enhancing reliability.

\subsection{Language-Based Geometric Reasoning}
\label{subsec:lan}

MLLMs have demonstrated high proficiency in handling mathematical and logic tasks~\cite{lewkowycz2022solving, bubeck2023sparks}. Recent benchmarking studies~\cite{zhang2024geoeval} show that MLLMs can effectively interpret textual problem statements and reason over both 2D and 3D geometry. This ability is particularly important for spatial reasoning tasks, which often require precise calculations of distance, direction, and other spatial relationships.
 
To leverage the language-based reasoning strengths of MLLMs, we represent the extracted 3D geometric information in a text-based format. In particular, each object is assigned a unique ID, and its geometric attributes are explicitly expressed as a text sequence. For a 3D scene, a list of such object geometric descriptions is provided as contextual input along with task-specific prompts. Thanks to the strong language understanding abilities of MLLMs, the exact form of text sequences can be flexible, allowing models to handle a mixture of different geometric primitives (see the example in~\cref{fig:3drep}). The level of geometric details can also be easily adjusted according to task requirements and data availability. While previous work also extracts object-level 3D information, it is typically used to organize input tokens~\cite{huang2024embodied, huang2024chat} or serve as auxiliary outputs~\cite{LLaVA3D2024}, where precisely defined geometric data are not fully utilized for reasoning. 

Another advantage of using languages for geometric reasoning is the enhanced explainability, which is mostly absent in smaller fine-tuned models. Modern MLLMs can generate chain-of-thought reasoning, in which intermediate steps are explicitly articulated. The step-by-step calculation and reasoning not only reveal how a final answer is derived but also facilitate error analysis and model refinement. By examining intermediate steps, it becomes possible to identify systematic mistakes, assess model reliability, and design targeted strategies~\cite{zhou2023leastto, wang2023selfconsistency} to reduce errors and enhance performance.

\subsection{Text-Image Association}
\label{subsec:im}

When provided with both images and geometric descriptions, MLLMs often struggle to associate the two modalities, particularly in complex scenes containing many objects. This challenge is evident in our experiments, as discussed in~\cref{subsubsec:abl}. To address this issue, we annotate objects on images with their IDs, which establishes a clear correspondence between visual and geometric information. Given an object with geometric attributes, we project its object center $C$ to the image by first transforming it into camera canonical coordinates: 
\begin{equation}
  [x,y,z]^T = R C + t,
  \label{eq:proj}
\end{equation}
and then calculating its pixel coordinates $(i, j)$ on the image plane with camera intrinsics $K$. At the projected location $(i, j)$, we place the object IDs. For each image, we annotate all the objects whose centers are mapped inside image bounds.

\begin{figure}[t]
  \centering
  \includegraphics[width=0.95\linewidth]{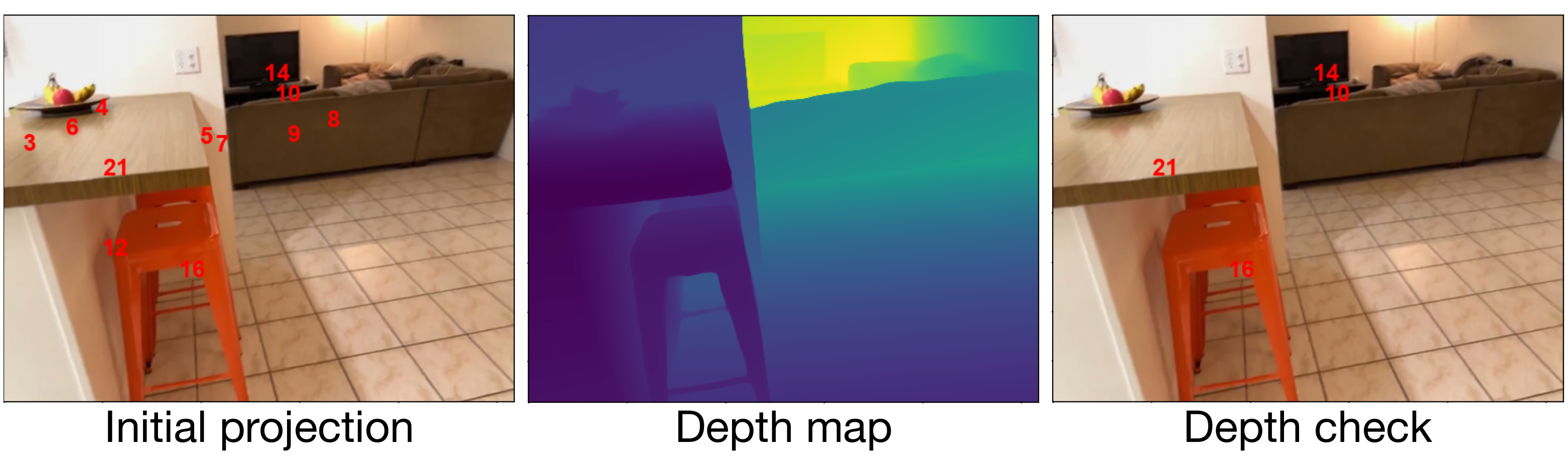}
  \caption{Object annotation with occlusion check. Left: initial object annotation through projection. Middle: depth map from 3D reconstruction. Right: object annotation after depth-based occlusion check.}
  \label{fig:occl}
 \end{figure}

During object center projection, no occlusion check is performed. As a result, occluded objects are also marked, leading to incorrect association. This issue is illustrated in the left image in~\cref{fig:occl}. While one solution is to render full geometric shapes for an occlusion check, it significantly increases computational cost. Instead, we exploit the depth maps $D$ that are output from 3D reconstruction models. Specifically, we do not annotate an object if $z$ in~\cref{eq:proj} is larger than $D_{(i, j)}$, \ie, the object center is farther than the depth at the same location. As can be seen in~\cref{fig:occl}, most annotations corresponding to occluded objects are removed.  

While certain semantic attributes (e.g., object category) can be obtained from 3D scene analysis, we deliberately exclude such semantic information from textual descriptions. This design encourages MLLMs to infer semantic information directly from images. As MLLMs are shown to be highly capable in visual understanding~\cite{team2023gemini, hurst2024gpt}, they can effectively extract task-relevant semantics. Furthermore, MLLMs dynamically attend to image regions relevant to a given task, which is more computationally efficient than precomputing and storing exhaustive semantic analysis.

It is highly challenging to extract 3D information for every object in a scene, regardless of the chosen algorithm. Recent research~\cite{yang2025thinking} shows that MLLMs exhibit a strong understanding of relative positions of objects in close proximity, but it degrades significantly as object distance increases. Since annotated objects are registered to a global 3D coordinate system, MLLMs can use those objects as spatial anchors and infer the likely positions of missing objects based on their spatial relationships with nearby annotated ones. Therefore, our method to link objects in images to their geometric information allows MLLMs to tolerate missing detections and perform more robust and complete spatial reasoning.

\section{Experiments}

\subsection{Benchmarks}
\label{subsubsec:data}

We evaluate our approach on VSI-Bench~\cite{yang2025thinking}, and MindCube~\cite{yin2025spatial}. VSI-Bench contains over 5,000 questions derived from 288 in-door videos. This benchmark assesses a broad spectrum of spatial understanding capabilities, including configurational tasks (object counting, relative distance, relative direction, and route planning), measurement estimation tasks (object size, room size, and absolute distance), and a spatiotemporal task of determining the order of appearance. MindCube focuses on spatial reasoning from limited input views. It contains questions on object spatial relationships, where each question is based on a scene captured by 2 to 4 views, with various camera movement types. 

\subsection{Implementation Details}
\label{subsubsec:impl}

We adopt the $\pi^3$ model~\cite{wang2025pi} for reconstruction, as it achieves high accuracy and can predict metric scale. For video input, It is applied to subsampled frames. We also run Mask2Former~\cite{cheng2021mask2former} trained on ADE20K to obtain semantic segmentation for each input image. The reconstructed point clouds are rotated to be axis-aligned using the up direction estimated from floor points and the dominant direction of horizontal room boundaries. 

Because the point clouds are reconstructed from images, they are noisier than those from depth sensors, making existing 3D segmentation models unreliable. Instead, we back-project 2D semantic labels to the point cloud and aggregate them using a voxel grid, assigning each voxel the majority label of its points. Spatially connected voxels with the same label are grouped into objects, and 3D bounding boxes are computed using their center coordinates and side lengths. For objects such as round tables, we fit vertical cylinders by projecting voxel groups onto the $xy$-plane and fitting circles using least squares, with height determined by the voxel group’s $z$ extent. Extracted geometric attributes are converted into compact textual references indexed by numeric IDs, and the IDs are placed on the images via camera projection, as described in~\cref{subsec:recnstr}.

\begin{figure}
  \centering
  \includegraphics[width=0.95\linewidth]{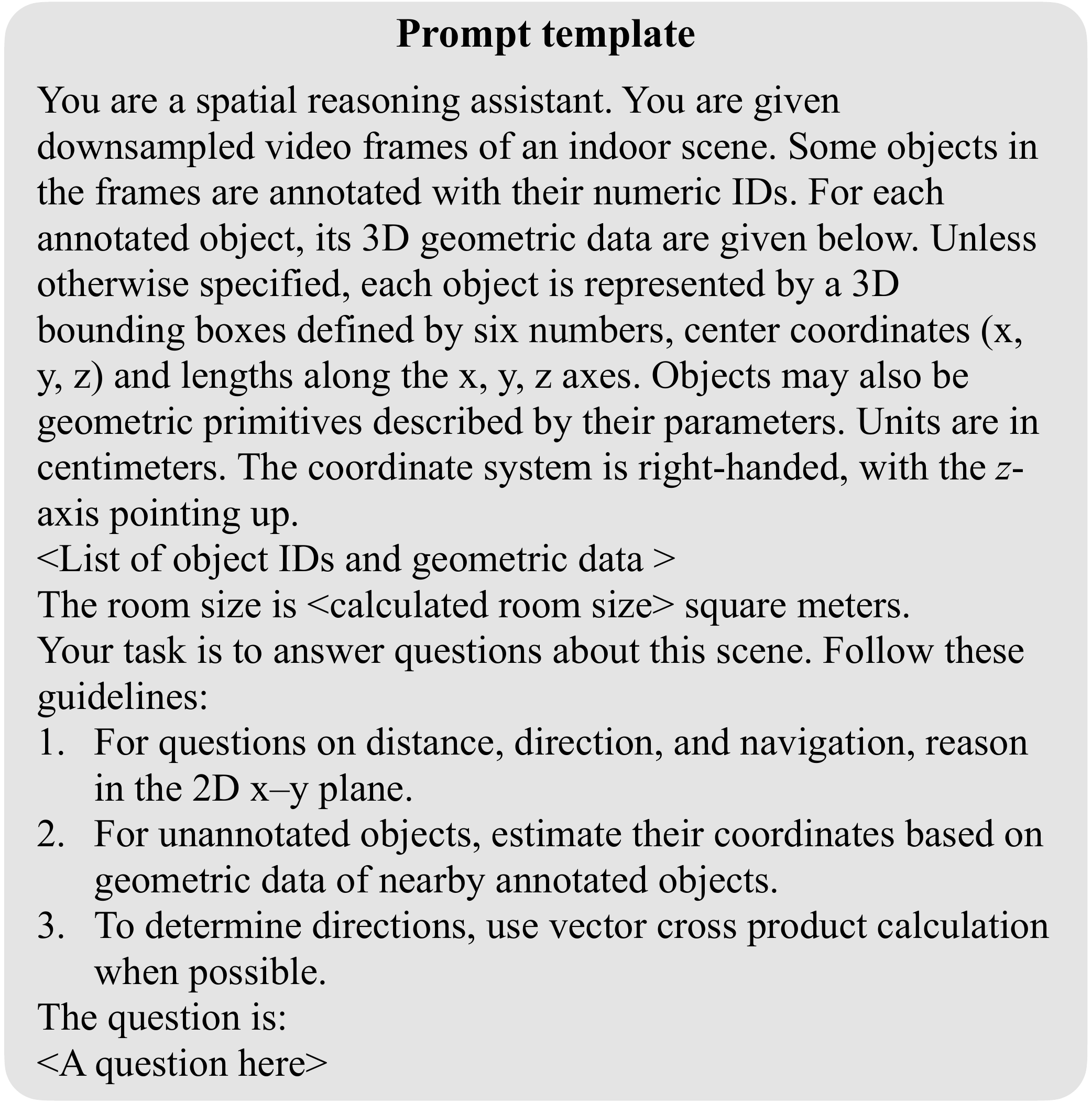}
  \caption{Prompt template used in evaluations.}
  \label{fig:prompt}
 \end{figure}

The GR3D input consists of annotated images and textual references inserted into a prompt template. Fig.~\ref{fig:prompt} shows the template used in the experiments, which explains the structure of the representation. We observe that it is important to specify the handedness and the upright orientation of the coordinate system, as this allows MLLMs to correctly apply orientation tests to solve direction-related questions. To reduce token usage, we provide the pre-computed area of room boundary polygons for room size questions. 

\subsection{Evaluation Results}
\label{subsubsec:res}

Evaluation results on VSI-Bench are summarized in~\cref{tab:perf}. For comparison, we include the benchmark scores of GPT-4o and Gemini-1.5 Pro, reported in~\cite{yang2025thinking}, as well as the scores of InternVL2~\cite{chen2024far}, a leading open-source mid-size model on this benchmark. VG LLM~\cite{zheng2025learning} is among the few models fine-tuned with 3D scene data and evaluated on VSI-Bench. In addition, we evaluate GPT-5 on this benchmark~\footnote{We set reasoning effort to low, which provides a favorable balance between accuracy and efficiency.}. To assess the effectiveness of our method, we measure the performance of incorporating GR3D with InternVL2, GPT-4o and GPT-5~\footnote{We follow the setting in~\cite{yang2025thinking}, using 32 frames for InternVL2 and 16 frames for the GPT models.}.

\begin{table*}
  \centering
  \small
  \begin{tabularx}{0.93\textwidth}{l|cccccccc|c}
    \toprule
    Method Name & \rotatebox{55}{Obj. Count} & \rotatebox{55}{Abs. Dist.} &  \rotatebox{55}{Obj. Size} &  \rotatebox{55}{Room Size} &  \rotatebox{55}{Rel. Dist.} &  \rotatebox{55}{Rel. Dir.} &  \rotatebox{55}{Route Plan} &  \rotatebox{55}{Appr. Order} & Avg. \\
    \midrule
    InternVL2-8B~\cite{chen2024far} & 31.3 & 29.0 & 48.9 & 44.2 & 38.0 & 33.4 & 28.9 & 46.4 & 37.5 \\ 
    VG LLM-8B~\cite{zheng2025learning} & \textbf{67.9} & 37.7 & 58.6 & 62.0 & 46.6 & 40.7 & 32.4 & 59.2 & 50.7 \\
    Gemini-1.5 Pro~\cite{team2024gemini} & 56.2 & 30.9 & 64.1 & 43.6 & \textbf{51.3} & 46.3 & 36.0 & 34.6 & 45.4 \\
    GPT-4o-2024-08-06~\cite{hurst2024gpt} & 46.2 & 5.3 & 43.8 & 38.2 & 37.0 & 41.3 & 31.5 & 28.5 & 34.0  \\
    GPT-5-2025-08-07~\cite{hurst2024gpt}  & 39.7 & 28.8 & \textbf{68.9} & 50.7 & 42.6 & 42.4 & 50.0 & \textbf{69.3} & 50.3 \\
     \midrule
    GR3D + InternVL2-8B & 33.9 & 29.3 & 46.2 & 63.7 & 37.1 & 40.3 & 30.1 & 53.7 & 42.5 \\ 
    GR3D + GPT-4o-2024-08-06  & 48.3 & 30.7& 44.1 & 66.3 & 42.1 & 40.7 & 46.4 & 44.7 & 45.3 \\
    GR3D + GPT-5-2025-08-07  &  52.2 & \textbf{48.2} & 62.5 & \textbf{72.1} & 50.8 & \textbf{69.5} & \textbf{58.7} & 66.2 & \textbf{59.8} \\
    \bottomrule
  \end{tabularx}
  \caption{VSI-Bench evaluation results. GR3D combined with GPT-5 achieves state-of-the-art performance. The best result for each task among all models is indicated in \textbf{bold}. }
  \label{tab:perf}
\end{table*}

Among all MLLMs, GPT-5 achieves the highest overall score. While the fine-tuned VG LLM attains a comparable performance, it should be noted that its training data include videos overlapping with VSI-Bench and questions similar to those in the benchmark. When integrated with GR3D, all three models exhibit performance gains, where models with stronger reasoning capabilities show greater improvements. In particular, GPT-5 achieves a 9\% increase, setting a new state of the art. This validates that our approach represents 3D spatial information in a form that can be effectively utilized by MLLMs. The improvements across all three models demonstrate that our approach can function as a general plug-in mechanism for enhancing spatial reasoning in MLLMs. As MLLMs continue to advance rapidly, our approach can immediately leverage these advancements without fine-tuning or retraining.

\cref{tab:mcperf} presents the MindCube results. We include models with leading performance reported in ~\cite{yin2025spatial}. Consistent with the trends observed on VSI-Bench, GR3D yields substantial performance improvements, boosting the accuracy of GPT-5 by 12\%. To our knowledge, this result represents the highest accuracy achieved by zero-shot methods on MindCube, highlighting the effectiveness of our representation in scenarios with sparse input views. Overall, the consistent gains across both VSI-Bench and MindCube demonstrate the generality of our approach and its ability to support diverse forms of 3D spatial reasoning without task-specific adaptation.

While our approach achieves strong overall performance, further analysis reveals that a large portion of the errors stem from inaccurate geometric attributes. In particular, our current implementation mainly relies on voxel connectivity to separate individual object instances, which struggles in cluttered scenes and thus reduces the accuracy of object counting and size estimation. More advanced 3D instance segmentation models could lead to improvements. In addition, when point clouds are highly noisy, the current object detection method tends to either miss objects or overestimate their dimensions. This issue can be mitigated through more robust 3D semantic segmentation or denoising techniques. 

\begin{table*}
  \centering
  \small
  \begin{tabularx}{0.55\textwidth}{l|ccc|c}
     \toprule
  Method Name & Rotation & Among & Around & Overall \\
  \midrule
  LLaVA-Onevision-7B~\cite{li2024llava}  & 36.5 & 48.4 & 44.1 & 47.4 \\
  DeepSeek-VL2-Small~\cite{lu2024deepseek}  & 37.0 & 50.4 & 26.9 & 47.6 \\
  GPT-5-2025-08-07~\cite{hurst2024gpt}  & 87.5 & 39.3 & 66.4 & 55.0 \\
  \midrule
  GR3D + GPT-5-2025-08-07  & \textbf{88.5} & \textbf{55.7} & \textbf{76.0} & \textbf{66.8} \\
   \bottomrule
    \end{tabularx}
  \caption{MindCube evaluation results. Rotation, among, and around represent subsets with different camera placements. GR3D combined with GPT-5 achieves the highest accuracy. The best result for each subset across all models is indicated in \textbf{bold}. }
  \label{tab:mcperf}
\end{table*}

\subsection{Ablation Study}
\label{subsubsec:abl}

In this section, we conduct ablation studies. We use VSI-Bench and GPT-5 for all experiments. Results are summarized in~\cref{tab:abl}. For brevity, we present scores of two configurational tasks and the average score of all tasks.   

\noindent\textbf{Effectiveness of object annotations.} We first investigate the role of object annotations by removing object IDs from images while keeping all other inputs unchanged. As shown in~\cref{tab:abl}, the resulting performance is nearly identical to GPT-5 alone. This indicates that without providing explicit association, MLLMs are unable to effectively utilize the provided geometric information.   

\noindent\textbf{Impact of reconstruction models.} To assess the impact of the reconstruction component, we replace $\pi^3$ with VGGT. As VGGT does not predict metric scales, we estimate a scaling factor from the ratio between the typical real-world heights and reconstructed heights of selected objects (e.g., ceilings, countertops, and desks). As can be seen, $\pi^3$ performs slightly better than VGGT, due to its improved reconstruction accuracy. We expect that continued advances in reconstruction models will further improve overall performance.

\noindent\textbf{Camera parameters for implicit association.} Instead of annotating objects in images via camera projection, we provide the full set of camera parameters for each input image, including focal length, the principal point, and the extrinsic matrix. These parameters allow MLLMs to compute the mapping between 3D space and the image plane when needed. We observe that the performance drops compared with the results above. By examining the reasoning process, we find that while GPT-5 can occasionally perform camera projection calculations correctly, its behavior is inconsistent and often error-prone, particularly for questions that require multiple projections.

\noindent\textbf{Textual scene descriptions with object ID citations.} Prior work~\cite{zhang2025pointvisiontext} reports that replacing raw images with text descriptions generated via image captioning can improve spatial reasoning performance. Motivated by this, we experiment with a similar approach in our setting. Specifically, we prompt GPT-5 to produce comprehensive and spatially detailed scene descriptions. In order to utilize geometric references, the captioning input is the images with object annotations, and the prompt requires that, whenever an annotated object is mentioned, its ID should be included next to the mention. This ensures that object mentions are tightly linked to their geometric attributes, similar to GR3D. An example of such scene descriptions is shown below: 

\begin{quote}
In the kitchen, a stainless steel trash bin [1] is positioned near the entrance. A microwave [2] sits above the counter. ...... A flat-screen TV [13] is mounted on the wall above a sleek wooden desk [12], which has a trash bin [14] underneath. A white standing fan [15] is placed near the window, ...... 
\end{quote}

As reported in~\cref{tab:abl}, this approach slightly outperforms GPT-5 alone, but it is inferior to GR3D. One possible reason is that the textual descriptions discard certain important visual details, particularly those related to local spatial relations. Nevertheless, this approach can be a lightweight alternative, especially in scenarios with a priority of data efficiency.

\begin{table}
  \centering
  \small
  \begin{tabularx}{0.94\linewidth}{l|cc|c}
    \toprule
    Method & Rel. Dir. & Route Plan & Avg. \\
    \midrule
    GR3D & 69.5 & 58.7 & 59.8 \\
    No object annotation & 51.3 & 52.0 & 50.3 \\
    VGGT reconstruction  & 62.2 & 56.7 & 58.3 \\
    Camera parameters &  42.6 & 45.4 & 48.8 \\
    Scene descriptions &51.9 & 51.0 & 53.4 \\
    \bottomrule
  \end{tabularx}
  \caption{Ablation study. We evaluate 1) removing object annotations in images, 2) replacing $\pi^3$ with VGGT, 3) providing camera parameters to implicitly associate images with object geometric information, and 4) converting images into scene descriptions with object ID citations. For brevity, we show the scores of two configurational tasks and the overall average. }
  \label{tab:abl}
\end{table}

\subsection{Qualitative Examples}
\label{subsec:sps}

\begin{figure*}
  \centering
  \includegraphics[width=0.97\linewidth]{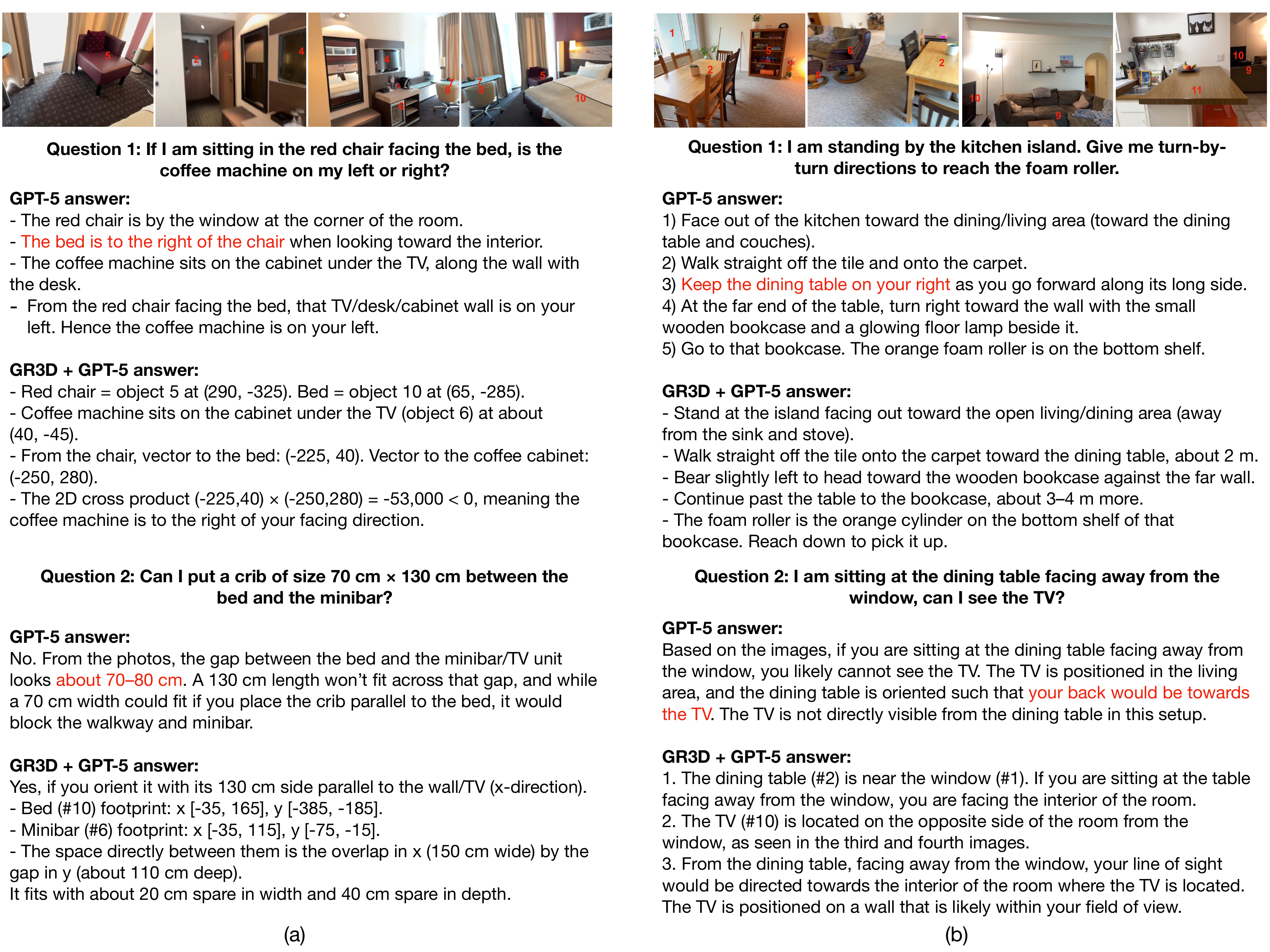}
  \caption{Example results of sparse view spatial reasoning. The part with incorrect reasoning is highlighted in {\color{red}red}. }
  \label{fig:sparsevw}
 \end{figure*}

To provide further insight on how GR3D improves spatial reasoning, we present qualitative examples that highlight its impact on model behavior in challenging cases.
In~\cref{fig:sparsevw}, we present example results from two scenes in VSI-Bench, each of which is captured by only four images subsampled from input videos. Four questions are posed, which require complex multi-step reasoning and an understanding of spatial context from an egocentric perspective. The questions are designed such that human can infer the answers from the images. We apply the same approach described in~\cref{subsubsec:impl}. For comparison, we also show the responses from GPT-5 alone, which receives the questions and unannotated images as input. Based on the results, we make the following observations: 

\begin{itemize}
\item Our method effectively exploits extracted geometric information to infer correct spatial relationships, whereas GPT-5 alone fails due to objects appearing in widely separated views. For example, in~\cref{fig:sparsevw}(a), our method first computes two vectors toward the bed and the cabinet and then calculates their cross product to determine directions. In contrast, in~\cref{fig:sparsevw}(b), GPT-5 mistakenly suggests walking on the left side of the table to reach the bookshelf and confuse the relative positions of the window and TV.
\item With GR3D, locations of unannotated objects (\eg, the coffee table and the foam roller) are correctly inferred based on the geometric information of nearby annotated objects (\eg., the cabinet and the bookshelf). 
\item Our approach leverages the advanced language reasoning capabilities of MLLMs to interpret complex spatial queries and generate transparent, step-by-step reasoning. Such a capability often lacks in smaller models. 
\item GR3D allows GPT-5 to perform 3D geometric reasoning while simultaneously leveraging its strong 2D visual understanding. For example, it can identify uncommon objects such as the foam roller.
 
\end{itemize}

\section{Conclusion}

We present a new approach to enhancing spatial reasoning in MLLMs through geometrically referenced 3D scene representations. By encoding object-level 3D attributes as structured textual references and explicitly linking them to annotated images, our method enables MLLMs to combine geometric reasoning with semantic perception in a transparent and interpretable manner. Our experiments on the VSI-Bench demonstrate that integrating GR3D with MLLMs substantially improves spatial reasoning and establishes a new state of the art in a zero-shot setting. These results highlight the value of bridging 2D semantics and 3D geometry via structured, language-accessible formats.

Our current pipeline consists of multiple components to extract geometric attributes, which are then converted into text-based descriptions. While this design allows each component to be flexibly replaced for improving performance, an interesting future direction is to move toward an end-to-end framework. Recent work such as SceneScript~\cite{avetisyan2024scenescript} demonstrates the feasibility of directly generating geometric information in form of structured text sequences from visual inputs. Extending this idea to our setting could result in a unified, learning-based model, potentially improving robustness and reducing error propagation across stages.

{
    \small
    \bibliographystyle{ieeenat_fullname}
    \bibliography{main}
}


\end{document}